# Using GPT-2 to Create Synthetic Data to Improve the Prediction Performance of NLP Machine Learning Classification Models

Dewayne Whitfield

December 31, 2020

## Abstract:
Classification Models use input data to predict the likelihood that the subsequent input data will fall into predetermined categories. To perform effective classifications, these models require large datasets for training. It is becoming common practice to utilize synthetic data to boost the performance of Machine Learning Models. It is reported that Shell is using synthetic data to build models to detect problems that rarely occur; for example Shell created synthetic data to help models to identify deteriorating oil lines.(Higginbotham, 2020) It is common practice for Machine Learning Practitioners to generate synthetic data by rotating, flipping, and cropping images to increase the volume of image data to train Convolutional Neural Networks(CNN). The purpose of this paper is to explore creating and utilizing synthetic NLP data to improve the performance of Natural Language Processing (NLP) Machine Learning Classification Models. In this paper I used a Yelp pizza restaurant reviews dataset and transfer learning to fine-tune a pre-trained GPT-2 Transformer Model to generate synthetic pizza reviews data. I then combined this synthetic data with the original genuine data to create a new joint dataset. For performance comparison purposes, I built two baseline models on two separate datasets using the Multinomial Naive Bayes Classifier algorithm. The two datasets were: The Yelp Pizza Reviews Dataset (450 observations) and a combined Yelp Pizza Reviews and Synthetic Yelp Reviews Dataset(11,380 observations). I created a separate ground truth dataset from the original Yelp Pizza Reviews Dataset that would serve as a ground truth test dataset. I then executed an analysis of the baseline models on the single ground truth test dataset to establish the following prediction performance metrics for each baseline model: precision, accuracy, recall, F1, and a confusion matrix. The combined Yelp Pizza Review Dataset outperformed the genuine Yelp Pizza Reviews Dataset on each of the performance metrics. The Baseline Model's Accuracy was increased from .8281 to .8913, a 7.63% increase. The Baseline Model's Precision was increased from .7979 to .8737, a 9.49% increase.

## 1. Business Value:
In 2018, U.S. companies spent nearly $19.2 Billion on data acquisition and solutions to manage, process, and analyze this data.(Sweeney, 2019) The technique discussed in this paper could reduce the cost of data acquisition by reducing the amount of data needed to train high performance NLP classification models. This technique could also be used to improve current

models by expanding training datasets. These benefits could lead to companies and organizations achieving their goals more effectively while minimizing costs.

## 2. Methodology:

### 2.1 Introduction
For the research conducted in this paper I used the GPT-2 transformer model and Yelp open dataset pizza reviews to create the synthetic data.

### 2.1 GPT-2
Developed by OpenAI, GPT-2 is a large-scale transformer-based language model that is pre-trained on a large corpus of text: 8 million high-quality webpages.(Cheng,2020) The objective of GPT-2 is to predict the next word given all of the previous words within some text(Radford, 2020). The GPT-2 model can be trained with an additional custom dataset using a method called transfer learning to produce more relevant text.

### 2.2 Yelp Open Dataset Reviews

| | stars | rating | text |
|---|---|---|---|
| 1 | 5 | Positive | first time eating there and everything was so yummy! great pizza and salad, my son had the meatball sub he said it was very good, must have been because he wouldn't share. highly recommend. |
| 2 | 5 | Positive | so i don't give out 5 stars frivolously, but this place deserves it. everything i had (summer rolls, vegan pizza burger and chocolate chip cookies) was delicious. the service was great and the waitress helped me narrow down the never ending yummy looking options on the menu. they even gave me a smoothie sample while i waited.\n\nnext time i'm in cleveland, i'll definitely stop back in. i may even be coming for dinner today. |
| 3 | 5 | Positive | my absolutely favorite mushroom pizza in the city, great brunch spot - really reasonable and you can always get a table, because it's really big and has indoor and outdoor space. i think the ice cream is homeade. a neighbourhood favorite! |
| 4 | 4 | Positive | the pizza is very, very good! we arrived 15 minutes before closing time and the restaurant staff were very accommodating! |
| 5 | 5 | Positive | this is probably my favorite bar in pittsburgh. the pizza is solid, the hot dogs are great, and the beer cave...mmmm. my only complaint is that they haven't opened another one on the other side of town! |

*Figure 1: Sample of Yelp Pizza Reviews Data Subset*

The Yelp Open Dataset contains anonymized reviews on various businesses and services (Yelp). For this paper I created a subset of data of pizza restaurant reviews. Within this subset of data I divided the ratings into "Positive" and "Negative". Ratings that were 4 or 5 stars were categorized as "Positive". Ratings that were 1 or 2 stars were categorized as "Negative". For this paper my Negative dataset contained 225 observations and the Positive dataset also contained 225 observations.

### 2.3 Technical Approach

The intent of the research in this paper is to train two GPT-2 models on a small subset of Positive and Negative Yelp Pizza Reviews data. I will then use the two GPT-2 models to produce synthetic Positive and Negative review datasets. I will finally combine the new synthetic datasets with the genuine dataset and fit a classification model to this dataset that will have the ability to determine negative and positive sentiment of pizza restaurant reviews.

### 2.3.1 Generating Synthetic Review Data

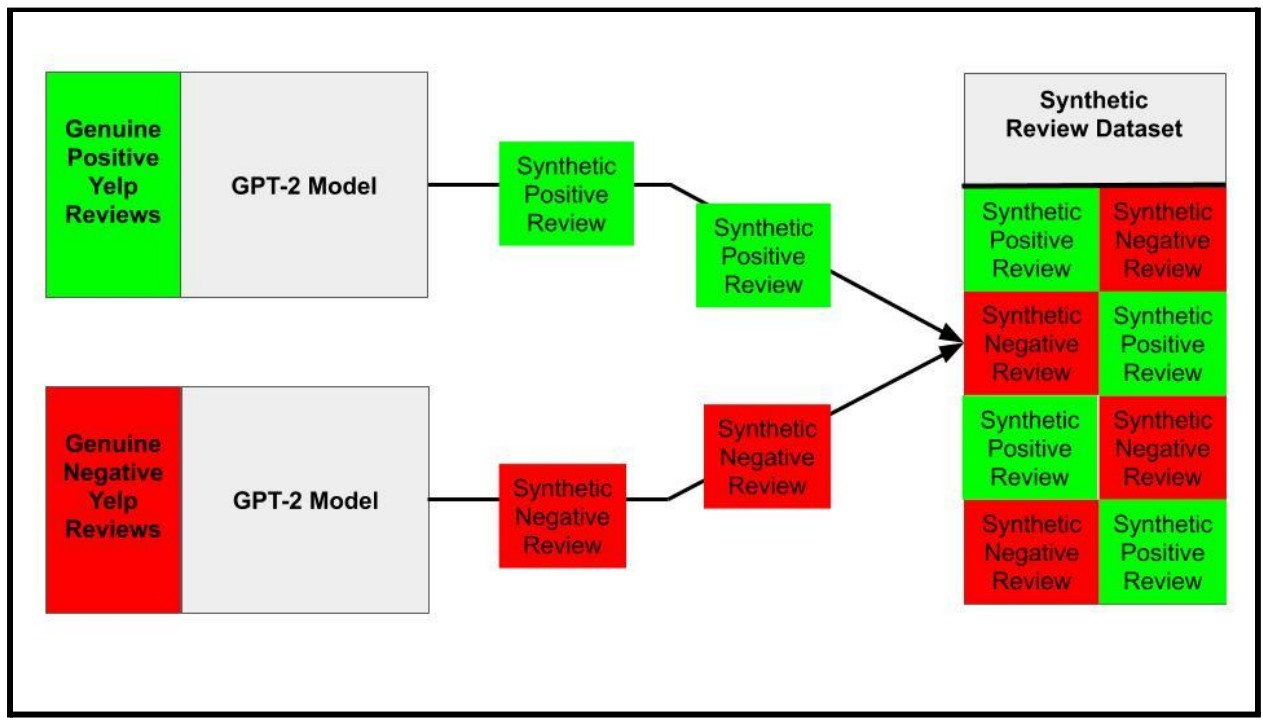

*Figure 2: Synthetic Review Generation and Dataflow*

My first task was to create two GPT-2 models and train one of them on genuine negative Yelp pizza review data and the other model on genuine positive Yelp pizza review data. I then had these models generate synthetic negative and positive review data that was combined into one a single dataset.

```
[ ]  #Negative
     sess = gpt2.start_tf_sess()

     gpt2.finetune(sess,
                   dataset=low_pizza_link,
                   model_name='355M',
                   steps=1000,
                   restore_from='fresh',
                   run_name='low-1',
                   print_every=10,
                   sample_every=200,
                   save_every=500
                   )
```

*Figure 3: GPT-2 Initiation*

## 2.3.2 Synthetic Review Generation

I chose the 355 million parameter GPT-2 model to build my two models. I used Google Colabs as my development notebook.

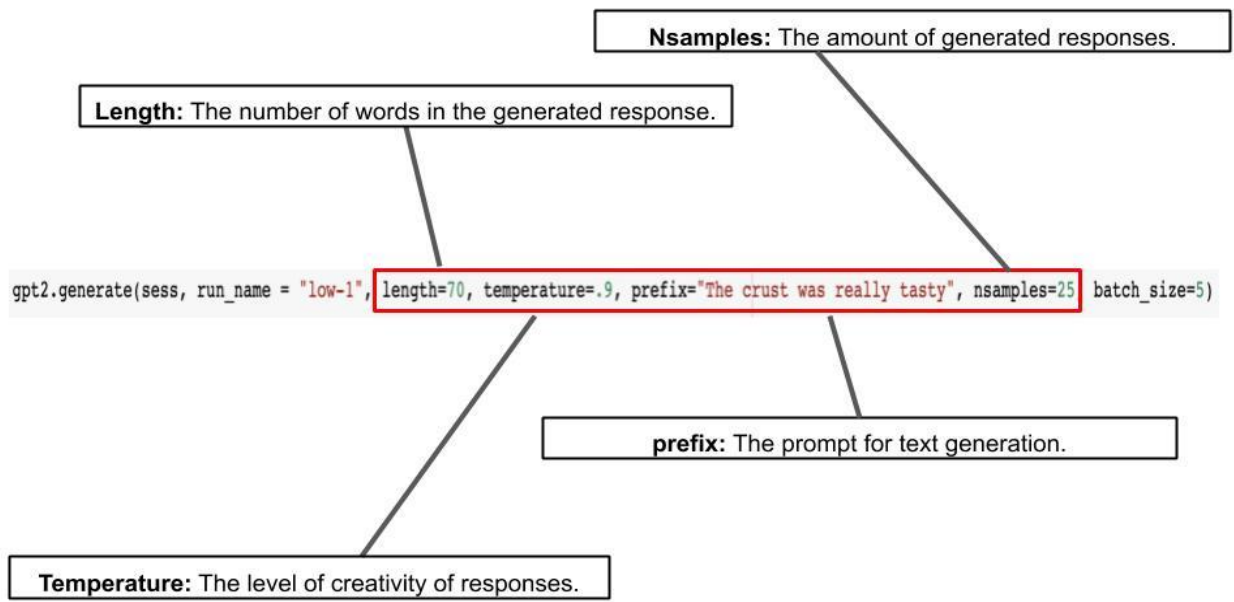

*Figure 4: GPT-2 text generation prompt*

I used the GPT2.Generate method to generate synthetic reviews. On average, my responses were 70 words long.

```
The crust was really tasty. i'd take it back again and i'd definitely recommend it to everyone. it was even tasty the next day!"
my favorite giordanos in az! the service and food is perfection. what i appreciate about it is that it's the closest tasting pizza compared to my ho
====================
The crust was really tasty. i had the garlic bread with powdered garlic and it was amazing. i had the eggplant florentine crudo pizza and it was oka
====================
The crust was really tasty. i loved the thin slices of pepperoni and the thick slices of meat. they were also very small, just enough for one slice.
cool place.  took my wife and son to lunch.  $20 for a really big slice of pizza, hot and sour.
====================
The crust was really tasty. i assume this applies to all of their pizza too. but the pineapples were the perfect size of chunks, plentiful, and did
====================
```

*Figure 5: Samples of GPT-2 Generated Prompts*

### 2.3.3 GPT-2 Text Generation

When generating synthetic reviews I wanted to ensure that the responses expanded on the genuine data and produced responses that were a strong representation of the genuine data. So when I wrote the prefix prompts I used words that were heavily represented in the genuine datasets. I wrote a Python Function that organized the genuine dataset corpus into trigrams (3 word consecutive combinations), bigrams (2 word combinations) and words. This function also provides a count and numerically sorts the occurrences of these words and combinations.

| Words | 80% - 100% | 50% - 80% | 25% - 50% |
|---|---|---|---|
| Trigrams | "Was very gross", "Bad customer service", "Pizza was cold" | "Not coming here", "Ingredients were stale", "Would not eat" | "Not very happy", "small portion size", "very salty taste" |
| Bigrams | "Soggy Crust", "Late delivery", "not happy" | "Dirty Table", "Money back", | "Very Greasy", "Never again", "was burned" |
| Words | "Pizza", "Angry", "Fresh", "Cheese" | "Incompetent", "Sauce", "Crust", "Nasty", "Delivery", "Service" | "Manager", "disgusting", "Stale" |

Prompt 1: Pizza was Nasty
Prompt 2: Not very happy with service
Prompt 3: Soggy Crust was very gross

*Table 1: Sample of GPT-2 Prompt Aid Tool*

### 2.3.4 Genuine and Synthetic Dataset Concatenation

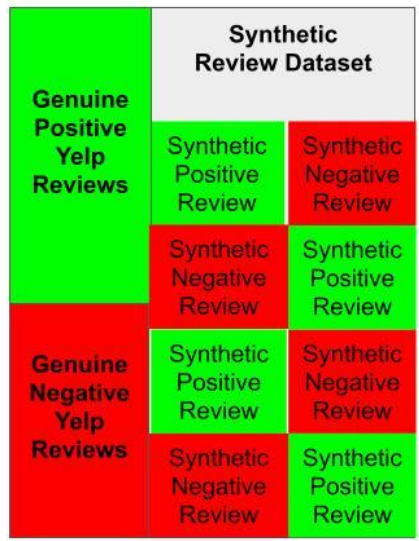

*Figure 6: Synthetic Data concatenated with Genuine Data*

After I created the synthetic Positive and Negative datasets I concatenated them with the genuine Negative and Positive Datasets.

## 2.3.5 Performance Testing

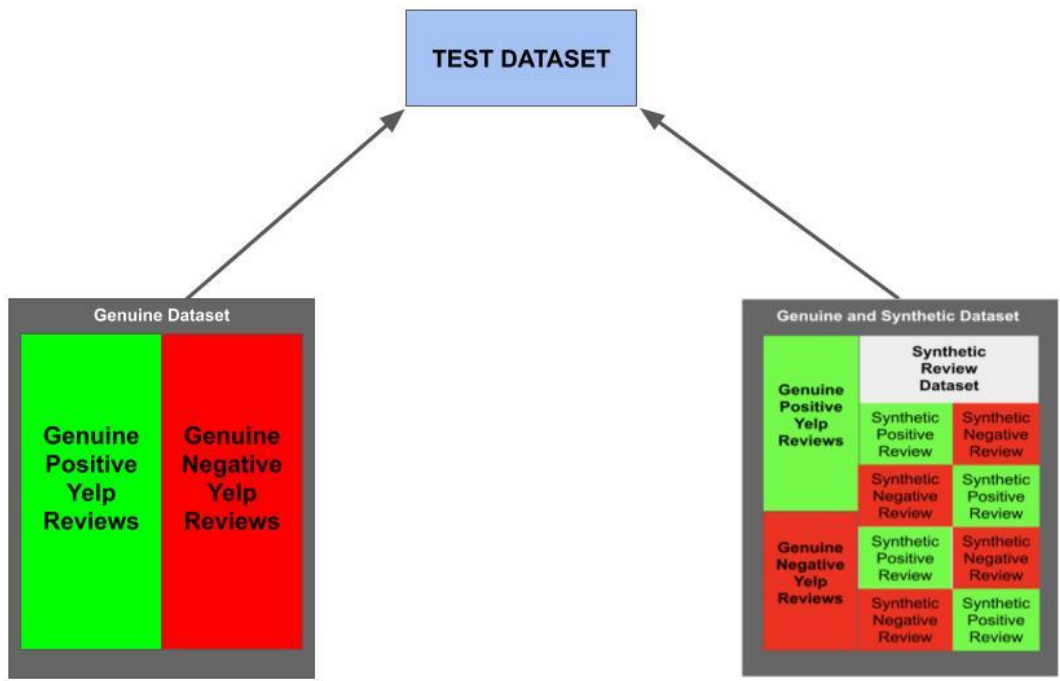

*Figure 7:  Baseline Model Testing*

To ensure there was a fair and equal analysis of the performance metrics, I used the scikit-learn train_test_split method to establish a single ground truth test set consisting of 198 observations derived from a totally separate dataset from the Yelp Open Dataset. I then built two baseline models on two datasets using the Multinomial Naive Bayes Classifier algorithm. The two datasets were: The genuine Yelp Pizza Reviews Dataset (450 observations) and the combined Genuine and Synthetic Yelp Reviews Dataset(11,380 observations).

## 2.3.6 Performance Testing Results

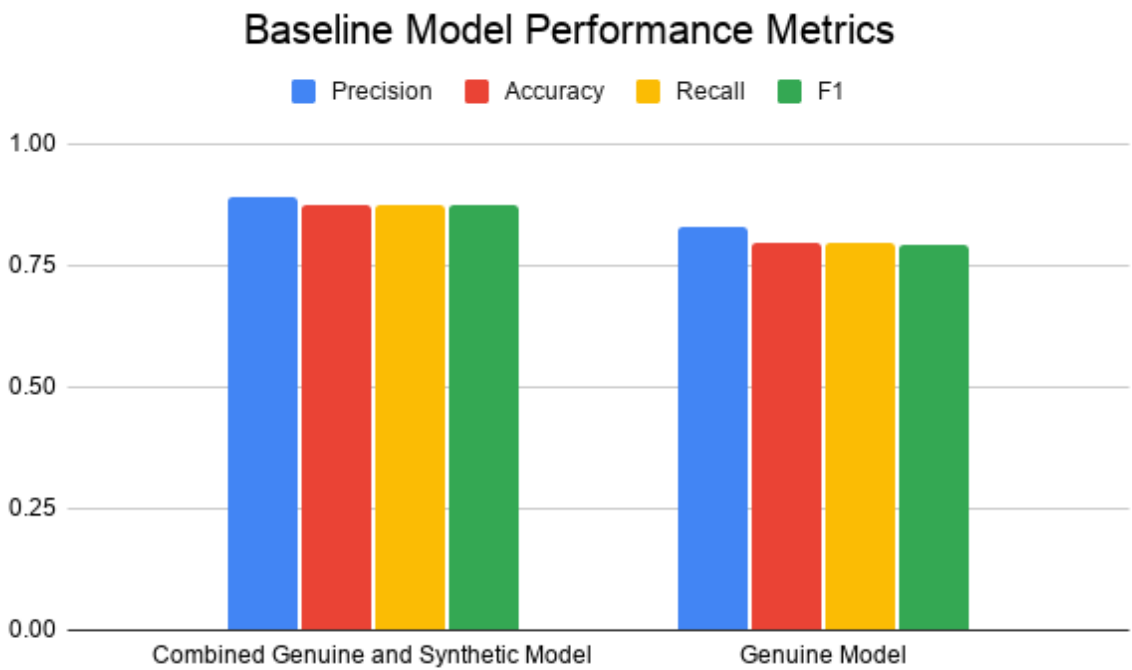

*Table 2: Baseline Model Performance Metrics*

I chose Precision, Accuracy, Recall, and F1 as performance metrics for the three baseline models. Overall, the Combined (Synthetic and Genuine) Model outperformed the Genuine Model on all performance metrics.

```
[[76 23]           [[94  5]
 [ 2 97]]           [35 64]]
```

**Synthetic and Genuine Model**     **Genuine Model**

*Figure 8: Baseline Model Confusion Model Analysis Results*

I also performed a Confusion Matrix Analysis. The Combined (Synthetic and Genuine) Model showed that the Genuine Model had more True Positive but less True Negatives when compared to the Combined (Synthetic and Genuine) Model. The Combined (Synthetic and

Genuine) Model had more False Positives but less False Negatives when compared to the Genuine Model.

## 3. Concluding Remarks:

In conclusion, the Synthetic and Genuine Dataset performed well in all performance metrics. This technique has the possibility of allowing organizations and businesses to build high performing NLP classification models without the high cost associated with large scale data acquisition. There are opportunities in exploring this technique on datasets with a larger observation count. There are also opportunities in exploring GPT-2 prompt design to better guide the GPT-2 model in generating relevant text. This is an exciting Machine Learning Technique that I feel deserves further exploration.

**Sources:**